\documentclass[10pt,twocolumn,letterpaper]{article}

\usepackage{cvpr}
\usepackage{times}
\usepackage{epsfig}
\usepackage{graphicx}
\usepackage{amsmath}
\usepackage{amssymb}
\usepackage{float} 
\usepackage{stfloats}
\usepackage{flafter}
\usepackage{blindtext}
\usepackage{graphicx}
\usepackage{authblk}
\usepackage[utf8]{inputenc}

\usepackage[breaklinks=true,bookmarks=false]{hyperref}

\cvprfinalcopy 

\def\cvprPaperID{****} 

\begin{document}

\title{Concurrence-Aware Long Short-Term Sub-Memories for \\Person-Person Action Recognition}

\author[$\dagger$ ]{Xiangbo Shu}
\affil[$\dagger$]{School of Computer Science and Engineering, Nanjing University of Science and
	Technology
}
\affil[ ]{\tt\small shuxb@njust.edu.cn}

\maketitle
\renewcommand{\thefootnote}{}
\footnotetext{$\ast$ Corresponding author.}
\renewcommand{\thefootnote}{\arabic{footnote}}
\begin{abstract}
	Recently, Long Short-Term Memory (LSTM) has become a popular choice to model individual dynamics for single-person action recognition due to its ability of modeling the temporal information in various ranges of dynamic contexts. However, existing RNN models only focus on capturing the temporal dynamics of the person-person interactions by naively combining the activity dynamics of individuals or modeling them as a whole. This neglects the inter-related dynamics of how person-person interactions change over time. To this end, we propose a novel Concurrence-Aware Long Short-Term Sub-Memories (Co-LSTSM) to model the long-term inter-related dynamics between two interacting people on the bounding boxes covering people. Specifically, for each frame, two sub-memory units store individual motion information, while a concurrent LSTM unit selectively integrates and stores inter-related motion information between interacting people from these two sub-memory units via a  new co-memory cell. Experimental results on the BIT and UT datasets show the superiority of Co-LSTSM compared with the state-of-the-art methods.

\end{abstract}

\section{Introduction}

\begin{figure}[t]
	\centering
	\vspace{3mm}
	\includegraphics[scale=0.35]{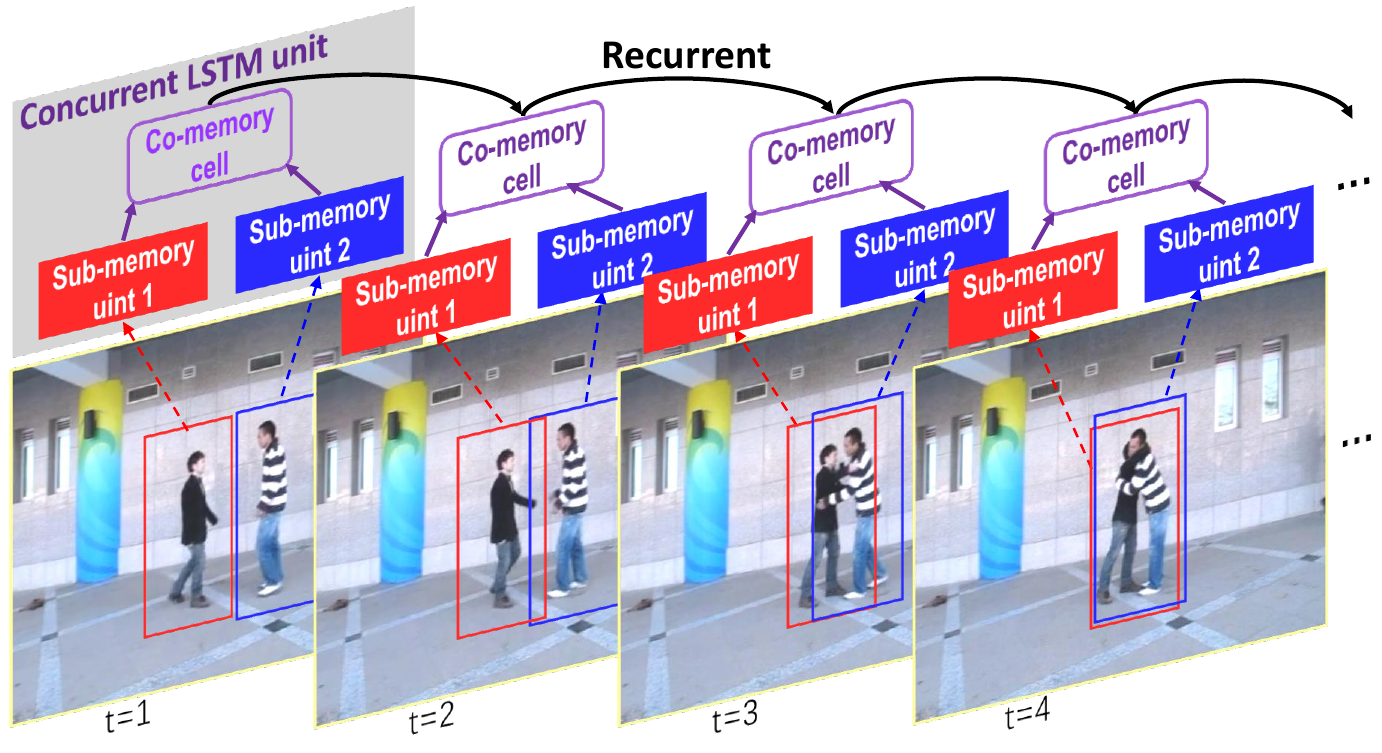}
	\caption{Illustration of the proposed Co-LSTSM. For each frame, two sub-memory units are developed to store individual motion information, while a {\bf concurrent} LSTM unit is developed to selectively integrate and store inter-related motion information between interacting people from two sub-memory units via a new co-memory cell ${\bf \pi}_t$ ($t=1,2,\cdots$). Stacked concurrent LSTM units are {\bf recurrent} to capture inter-related dynamics between interacting people over time. }\label{fig_idea}
\end{figure}

Person-person interaction (e.g., handshake, hug, etc), as the basic unit in the human activity, is attracting much attention in the computer vision and pattern recognition communities~\cite{kong2014interactive,kong2012leraning,chang2015learning,shariat2013a}. During a person-person interaction process, there are usually two individual motions from two interacting people respectively, some of which are concurrently inter-related with each other (e.g., two interacting people are stretching out hands in hug interaction). It has been proven that the concurrently inter-related motions between interacting people are discriminative for recognizing the person-person interactions~\cite{chang2015learning,kong2016close}. In most cases of person-person interaction, the concurrently inter-related motions between two interacting people are either 1) quite
symmetrically similar to each other (e.g., two interacting people are handshaking); or
2) not quite similar but are strongly interacting to each other (e.g., person A kicks person B, while person B retreats back). 

There are mainly two types of solutions for person-person interaction recognition. One solution (e.g., \cite{kong2014interactive,kong2012leraning,chang2015learning,zhang2012spatio}) is to extract the individual motion descriptors (e.g., spatio-temporal  interest
points~\cite{DollarVSPETS05cuboids}) from interacting people, and then predict the class label of an interaction by inferring the coherence between two individual motions. However, this solution regards the person-person interactions as two single-person actions, which ignores some inter-related motion information and brings in some irrelevant individual motion information. The other solution is to extract the motion descriptors on the interactive regions, and then train an interaction recognition model~\cite{kong2016close}. However, it is hard to locate interactive region before close interacting.

Usually, the difference between person-person interactions (e.g., boxing interaction and pat interaction) is subtle~\cite{ryoo2009spatio,chang2015learning,raptis2013poselet}, which brings in the challenge to recognize person-person interaction.
Recently, due to the powerful ability of capturing the sequential motion information, Recurrent Neural Networks (RNN)~\cite{williams1989learning,zhao2022wavelet,zhang2021skip}, especially Long Short-Term Memory (LSTM)~\cite{hochreiter1997long,fa2018global,wang2022spatiotemporal}, has proven successful on human action recognition tasks~\cite{donahue2015long,du2015hierarchical,veeriah2015differential,liu2016spatio,ibrahim2015hierarchical}. To well address the problem of person-person interaction recognition, we aim to explore the long-term inter-related dynamics between two interacting people by leveraging state-of-the-art LSTM model. However, existing LSTM models only modeling human individual dynamics independently do not consider the concurrently inter-related dynamics between interacting people. A naive way is to either 1) merge the individual actions at preprocessing stage~\cite{ke2016spatial} (e.g., consider interacting people as a whole); or 2) utilize two LSTM networks to model the individual dynamics of each interacting person respectively, and then fuse the output sequences from two LSTM networks~\cite{ibrahim2015hierarchical}. 
However, this neglects the inter-related dynamics between interacting people of how person-person interactions can change over time.

To this end, we propose a novel Concurrence-Aware Long Short-Term Sub-Memories (Co-LSTSM) for person-person interaction recognition by modeling the long-term inter-related dynamics between two interacting people on the bounding boxes covering people. It has the ability to aggregate the inter-related memories from individual memories of interacting people over time, as shown in Figure~\ref{fig_idea}. Specifically, we present a novel {concurrent} LSTM unit consisting of two sub-memory units that store the individual motion information on the bounding box covering people of each video frame. Following these two sub-memory units, a new co-memory cell selectively integrates and stores the memories from two sub-memory units to reveal the concurrently inter-related motion information between interacting people.
Overall, two interacting people in each frame are jointly modeled by a concurrent LSTM unit on the bounding boxes covering people, which outputs the concurrently inter-related hidden representations between interacting people rather than the individual hidden representations from individual human. The stacked concurrent LSTM units are {recurrent} in a time sequence to capture the concurrently inter-related dynamics between two interacting people over time. Extensive experiments on the widely-used benchmarks well show the superior performance of the proposed Co-LSTSM compared with the state-of-the-art methods and several baselines.

Our {\bf main contributions} in this work are two-fold: (1)~We propose a novel Concurrence-Aware Long Short-Term Memories (Co-LSTSM) to effectively address the problem of person-person interaction recognition. (2)~To our best knowledge, our work is the first attempt in modeling concurrently long-term inter-related dynamics over time between multiple motion objects by the variants of LSTM. 

\section{Related Work}
\subsection{Human Action Recognition}

Human activity recognition aims to automatically
understand the activities performed by people~\cite{chang2015learning,DBLP:conf/eccv/AmerXZTZ12,poppe2010survey},
including group-person interaction recognition (e.g., walking, queueing, etc)~\cite{lan2012discriminative,DBLP:conf/cvpr/RyooA06,choi2012unified,vahdat2011discriminative}, person-object interaction recognition (e.g., some people are eating, while the other people are riding a bike)~\cite{DBLP:conf/iccv/AmerTFZ13,DBLP:conf/eccv/AmerXZTZ12}, and person-person interaction recognition~\cite{kong2014interactive,kong2012leraning,chang2015learning,shariat2013a}.  
%

For group-person interaction recognition, one  solution used in~\cite{lan2012discriminative,DBLP:conf/cvpr/RyooA06} is to exploit the spatial distribution of human activities and present the spatio-temporal descriptors in
capturing the spatial distribution of people.  The other solution used   in~\cite{choi2012unified,patron2012structured,vahdat2011discriminative} is to  track all  body parts in a video, and then learn the holistic representations to estimate their collective activities.
In particular, instead of treating the two problems (i.e., tracking multiple people and estimating their collective activities) separately, Choi {\em et al.}~\cite{choi2012unified}  presented a unified framework to simultaneously track people and estimate their collective activities. Besides, Lan {\em et al.}~\cite{lan2010beyond,lan2012discriminative} proposed to recognize the group-person activities by jointly capturing the group activity, the individual human actions, and the interactions among them.

For person-object interaction recognition,  there are usually a number of concurrent individual activities (e.g., some people are riding a bike) and group activities (e.g., some people are walking together). To address this challenge, Amer {\em et al.}~\cite{DBLP:conf/iccv/AmerTFZ13} proposed a spatio-temporal AND-OR graph to jointly model the activity parts, person-person spatio-temporal relations, and person-object context, as well as enable multi-target tracking. Subsequently, Amer {\em et al.}~\cite{DBLP:conf/eccv/AmerXZTZ12} used a three-layered AND-OR graph to jointly model group activities, individual actions, and participating objects. A key point is that these methods require a multitude of detectors at different levels.

For person-person interaction recognition, some representative works~\cite{kong2014interactive,kong2012leraning,zhang2012spatio} used several interactive phrases as the latent mid-level feature to infer the person-person interaction from the human individual actions. Interactive phrases incorporating rich human knowledge provide an effective way to represent person-person interactions. However, the difference of some interactions (e.g., boxing and pat) is too stable to be discriminated only by the interactive phrases. Besides, some person-person interactions are complex, which cannot be described well by a certain amount of interactive phrases. Recently, Kong {\em et al.}~\cite{kong2016close} developed a patch-aware latent SVM to recognize the  interactions  by  inferring  the closely interactive regions between interacting people. However, it is hard to capture the interactive regions before close interacting. Moreover, Chang {\em et al.}~\cite{chang2015learning} proposed to extract features of each interacting person and then learn an interaction matrix between interacting people.

\subsection{RNN-based Action Recognition}

As neural nets for handling sequential data with variable length, RNN, especially LSTM, has been successfully applied to action recognition~\cite{zhu2016co,donahue2015long,du2015hierarchical,veeriah2015differential,liu2016spatio,ibrahim2015hierarchical}.
Many RNN-based action recognition methods are embedded the LSTM layer into Convolutional Neural Networks (CNN)~\cite{donahue2015long,wu2016multi,ke2016spatial,tang2019coherence,liu2017face,lai2016instance,shu2022expansion,wang2019region,li2021storyboard}.  For example,
Wu {\em et al.}~\cite{wu2016multi} proposed to train three types of CNNs equipped with LSTM to model the spatial, short-term motion and audio clues corresponding to the inputs of video frames, stacked optical flows, and audio spectrogram, respectively. Besides, some skeleton-based action recognition methods utilized RNN to model the long-term contextual
information of all skeletons. For example. Du {\em et al.}~\cite{du2015hierarchical} proposed a multilayer RNN framework to feed the five body parts from human skeletons into five subnets. As the number of layers increases, the representations outputs from several subnets are hierarchically fused to the inputs of the higher layers.

Some works aim to design the specific RNN architecture for the different action recognition tasks~\cite{ibrahim2015hierarchical,zhu2016co,shahroudy2016ntu,shu2020host,shu2021spatiotemporal,hu2019semantic,yan2020higcin,yan2018participation,yan2020social,yan2021position}. For example, in order to capture the co-occurrences
of discriminative joints, Zhu {\em et al.}~\cite{zhu2016co} added a mixed-norm regularization penalty to the deep LSTM networks. Moreover, the authors proposed an internal dropout technique to c
operate on the gates, cells, and output responses of the
LSTM nodes. To emphasize on the temporal change of motion information between two consecutive frames with the time, Veeriah {\em et al.}~\cite{veeriah2015differential} proposed a Differential RNN architecture equipped with the Derivative of States between the LSTM gates. Recently, Shahroudy {\em et al.}~\cite{shahroudy2016ntu} proposed
a Part-aware LSTM that separates the memory cell into the several sub-cells corresponding to the different body parts and explicitly models
the dependencies over spatial
and temporal domains concurrently. 
Likewise, Liu {\em et al.}~\cite{liu2016spatio} also proposed the similar LSTM architecture by pushing the traditional LSTM-based learning into temporal domains and spatial domains simultaneously.

Unlike existing RNN-based action recognition works, we consider the more challenging action recognition scenario within person-person interactions. To capture the interactive motion information rather than the individual motion information, the proposed Co-LSTSM explicitly models the concurrently inter-related dynamics between interacting people. The most related works~\cite{ke2016spatial,ibrahim2015hierarchical} either combine the individual dynamics of each person or treat the two interacting people as a whole. To our best knowledge, our work is the first time to model the concurrently long-term inter-related dynamics over time between interacting people by the LSTM-based model.

\section{Preliminary: RNN for Individual Action}
Given an input video clip $\{{\bf x}_t\in \mathbb{R}^n|t=1,\cdots,T\}$ with the length $T$, RNN~\cite{williams1989learning} models its dynamics through a  sequence of hidden states
$\{{\bf h}_t\in \mathbb{R}^m|t=1,\cdots,T\}$ with $M$ hidden units, which can be mapped to an output sequence $\{{\bf z}_t\in \mathbb{R}^k|t=1,\cdots,T\}$ ($k$ is the number of the classes of actions), i.e.,
\begin{equation} \label{eq1}
	\begin{aligned}
		{\bf{h}}_t = \varphi ({\bf{W}}_{hx} \cdot {\bf{x}}_t + {\bf{W}}_{hh} \cdot {{\bf{h}}_{t - 1}} + {\bf{b}}_h);
	\end{aligned}
\end{equation}
\begin{equation} \label{eq2}
	\begin{aligned}
		{\bf{z}}_t = \varphi ({\bf{W}}_{zh} \cdot {\bf{h}}_t  + {\bf{b}}_z),
	\end{aligned}
\end{equation}
where $\varphi(\cdot)$ denotes $tanh(\cdot)$, ${\bf{W}}_{h*}$ and ${\bf{W}}_{z*}$ are the weight matrices, and ${\bf{b}}_*$ is the bias vector. Finally, the output ${\bf z}_t$ at time step $t$ can be solved by a softmax function, i.e., ${{{y}}_{t,l}} = {{\exp ({z_{t,l}})}}/{{\sum\limits_{j = 1} {\exp ({z_{t,j}})} }}$, where the $j$-th element $z_{t,j}$ denotes the encoding of the confidence score on the $j$-th class action.

Due to  the exponential decay in retaining the context information of video frames, Long Short-Term Memory~\cite{hochreiter1997long}, a variant of RNN, provides a solution by allowing the network to learn when to forget previous hidden states and when to update hidden states given new information~\cite{donahue2015long}. 

Usually, each LSTM unit contains a memory cell (denoted by ${\bf c }_t$) storing the memory of the input sequence up to the time step $t$. In order  to store
the memory w.r.t the motion information in the long time, three types of gates (i.e., input gate ${\bf i}_t$, forget gate ${\bf f}_t$ and output gate ${\bf o}_t$) are incorporated into the LSTM unit to
control what information would enter and leave the memory cell over time~\cite{hochreiter1997long}, 
as follows,
\begin{equation} \label{eq4}
	\begin{aligned}
		{\bf{i}}_t = \sigma ({\bf{W}}_{ix} \cdot {\bf{x}}_t + {\bf{W}}_{ih} \cdot {{\bf{h}}_{t - 1}} + {\bf{b}}_i);
	\end{aligned}
\end{equation}
\begin{equation} \label{eq5}
	\begin{aligned}
		{\bf{f}}_t = \sigma ({\bf{W}}_{fx} \cdot {\bf{x}}_t + {\bf{W}}_{fh} \cdot {{\bf{h}}_{t - 1}} + {\bf{b}}_f);
	\end{aligned}
\end{equation}
\begin{equation} \label{eq6}
	\begin{aligned}
		{{\bf{o}}_t} = \sigma ({\bf{W}}_{o{{x}}}^{} \cdot{\bf{x}}_t  + {\bf{W}}_{oh} \cdot {{\bf{h}}_{t - 1}} + {\bf{b}}_o),
	\end{aligned}
\end{equation}
where $\sigma(\cdot)$ is a sigmoid function; 
${\bf{W}}_{*x}$ and ${\bf{W}}_{*h}$ are the weight matrices; ${\bf b}_*$ is the bias vector. In addition to three gates, the memory cell ${\bf c}_t$ can be expressed as
\begin{equation} \label{eq7}
	\begin{aligned}
		{\bf{c}}_t = {\bf{f}}_t^s \odot {\bf{c}}_{t - 1} + {\bf{i}}_t \odot {\bf{g}}_t,
	\end{aligned}
\end{equation}
where ${\bf{g}}_t = \varphi ({\bf{W}}_{gx} \cdot {\bf{x}}_t + {\bf{W}}_{gh} \cdot {{\bf{h}}_{t - 1}} + {\bf{b}}_g)$, and $\odot$ denotes the element-wise product.
Finally, a hidden state ${{\bf{h}}_t}$ at time step $t$ can be expressed as
\begin{equation} \label{eq9}
	\begin{aligned}
		{{\bf{h}}_t} = {{\bf{o}}_t} \odot \varphi ({{\bf{c }}_t}).
	\end{aligned}
\end{equation}


\section{The Proposed Co-LSTSM}
\subsection{The Architecture}
For  person-person  interaction recognition,  each video frame
contain two concurrent individual actions from interacting people, some of which are inter-related with each other. Existing LSTM models targeting to singe-person actions cannot handle the person-person interactions well. As mentioned before, we can roughly treat two interacting people as a whole before training the LSTM network. However, this solution will bring in some individual-special motion information. Besides, we can also model the individual dynamics of each person by two LSTM networks respectively, and then naively combine (e.g., concatenate or pool) the output sequences from two LSTM networks into the final representation. However, it is intuitive that this strategy loses some concurrently inter-related motion information between interacting people.
\begin{figure}[t]
	\centering
	\includegraphics[scale=0.32]{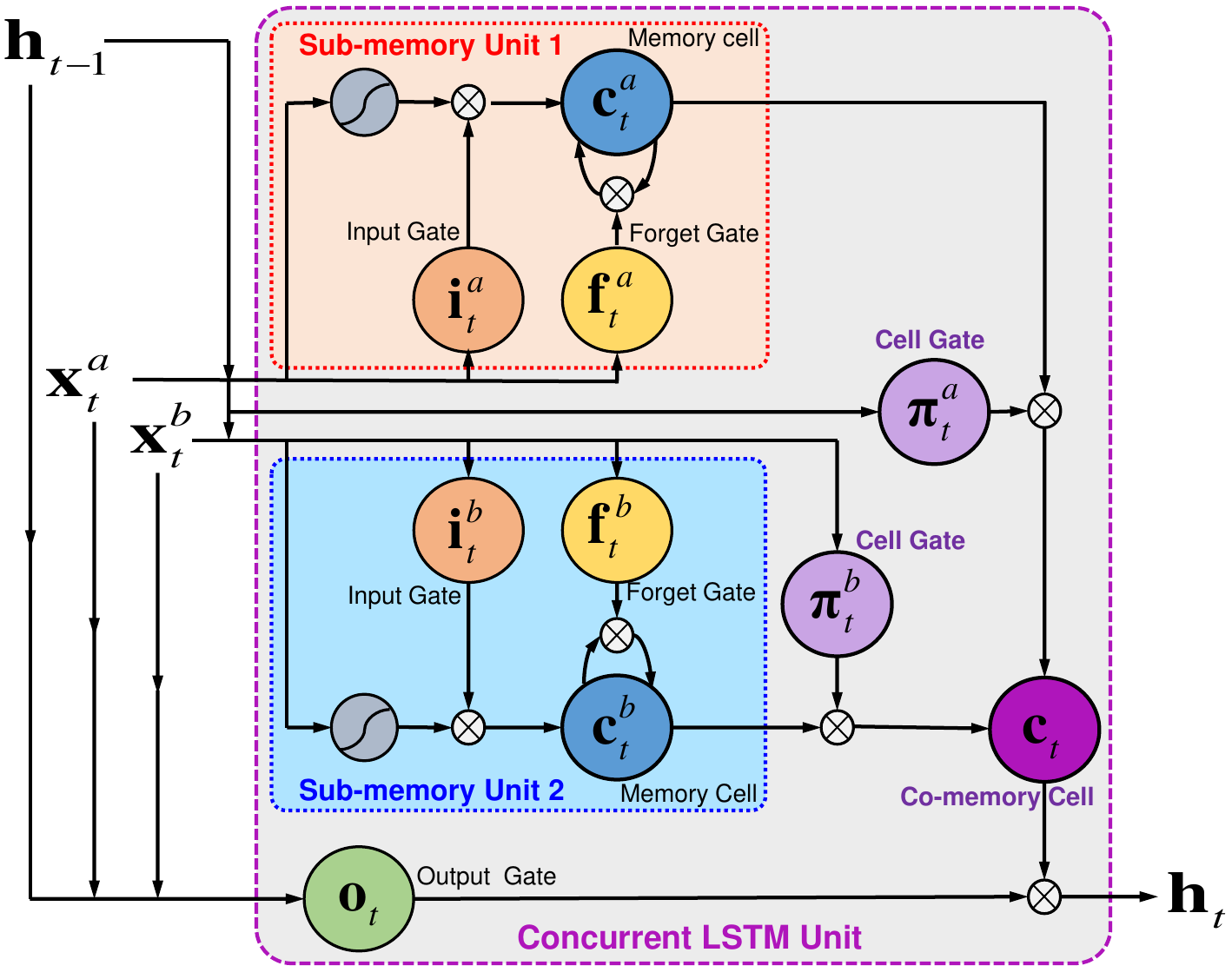}
	\caption{ Illustration of a concurrent LSTM unit in the proposed Co-LSTSM. For the concurrent inputs ${\bf x}_t^a$ and ${\bf x}_t^b$ at time step $t$, a concurrent LSTM unit consists of two specific sub-memory units, a common output gate ${\bf o}_t$, two new cell gates (i.e., ${\bf \pi}_t^a$ and ${\bf \pi}_t^b$) and a new co-memory cell ${\bf c}_t$. These two sub-memory units includes the respective input gates (i.e., ${\bf i}_t^a$ and ${\bf i}_t^b$), forget gates (i.e., ${\bf f}_t^a$ and ${\bf f}_t^b$), sub-memory cells (i.e., ${\bf c}_t^a$ and ${\bf c}_t^b$). In particular, two sub-memory cells (i.e., ${\bf c}_t^a$ and ${\bf c}_t^b$) are jointly fed into the co-memory cell  ${\bf c}_t$, followed by the hidden representation ${\bf h}_t$.}
	\label{fig_crnn}
\end{figure}

To this end, we propose a Concurrence-Aware Long Short-Term Memories (Co-LSTSM) to capture the concurrently inter-related dynamics between interacting people rather than the individual dynamics of each person. Our key idea is to develop two sub-memory units to store the individual motion information of each person respectively, and a concurrent LSTM unit to selectively integrate and store the concurrently inter-related motion information between interacting people from the individual motion information.  Figure~\ref{fig_crnn} illustrates the architecture of a concurrent LSTM unit of the proposed Co-LSTSM. Overall, the concurrent LSTM unit at each time step consists of two specific sub-memory units, two cell gates, a common output gate and a new co-memory cell. Specifically, these two sub-memory units include their respective input gates, forget gates, memory cells. And the co-memory cell between two sub-memory units selectively integrates the individual motion information from two memory units and memorizes the inter-related motion information.

Formally,  $\{{\bf x}_t^a\in \mathbb{R}^n|t=1,\cdots,T\}$ and $\{{\bf x}_t^b\in \mathbb{R}^n|t=1,\cdots,T\}$ denote two sequences of two concurrent people, respectively; ${\bf{i}}_t^a$, ${\bf{f}}_t^a$ and  ${\bf{c}}_t^a$ denote the input gate, forget gate and sub-memory cell in sub-memory unit 1 at time step $t$, respectively; ${\bf{i}}_t^b$, ${\bf{f}}_t^b$ and  ${\bf{c}}_t^b$ denote the input gate, forget gate and sub-memory cell in sub-memory unit 2 at time step $t$, respectively. All of them can be expressed in the following equations
\begin{equation} \label{eq10}
	\begin{aligned}
		{\bf{i}}_t^s = \sigma ({\bf{W}}_{ix}^s \cdot {\bf{x}}_t^s + {\bf{W}}_{ih}^s \cdot {{\bf{h}}_{t - 1}} + {\bf{b}}_i^s), s\in \{a,b\};
	\end{aligned}
\end{equation}
\begin{equation} \label{eq11}
	\begin{aligned}
		{\bf{f}}_t^s = \sigma ({\bf{W}}_{fx}^s \cdot {\bf{x}}_t^s + {\bf{W}}_{fh}^s \cdot {{\bf{h}}_{t - 1}} + {\bf{b}}_f^s), s\in \{a,b\};
	\end{aligned}
\end{equation}
\begin{equation} \label{eq12}
	\begin{aligned}
		{\bf{g}}_t^s = \varphi ({\bf{W}}_{gx}^s \cdot {\bf{x}}_t^s + {\bf{W}}_{gh}^s \cdot {{\bf{h}}_{t - 1}} + {\bf{b}}_g^s), s\in \{a,b\};
	\end{aligned}
\end{equation}
\begin{equation} \label{eq13}
	\begin{aligned}
		{\bf{c}}_t^s = {\bf{f}}_t^s \odot {\bf{c}}_{t - 1}^s + {\bf{i}}_t^s \odot {\bf{g}}_t^s, s\in \{a,b\},
	\end{aligned}
\end{equation}
where $W_{*x}^s$ and $W_{*h}^s$ are the weight matrices, and $b_*$ is the bias vector.

Two cell gates ${\bf \pi}_t^a$ and  ${\bf \pi}_t^b$ following the sub-memory unit 1 and the sub-memory unit 1 respectively aim to control what memories from two sub-memory units enter and leave at each time step. Unlike the traditional gates, the cell gate  ${\bf \pi}_t^s$ ($s\in \{a,b\}$) is activated by a nonlinear function of two inputs ${\bf x}_t^a$ and ${\bf x}_t^b$ and the past hidden state ${\bf h}_{t-1}$, i.e.,
\begin{equation} \label{eq13.5}
	\begin{aligned}
		{{\bf{\pi}}_t^s} = \sigma ({\bf{W}}_{\pi{\rm{x}}}^{s} \cdot 
				{{\bf{x}}_t^s} + {\bf{W}}_{\pi h}^{} \cdot {{\bf{h}}_{t - 1}} + {\bf{b}}_\pi^{}), s\in \{a,b\},		\end{aligned}
	\end{equation}
	where $s\in \{a,b\}$, $W_{\pi *}$ are the weight matrices, and $b_{\pi}$ is the bias vector.
	Based on the consistent interactions between two interacting people, these two cell gates ${\bf \pi}_t^s$ ($s\in \{a,b\}$) allow more concurrently inter-related motion information between interacting people to enter the co-memory cell ${\bf c}_t$ and contribute to one common hidden state. In this work, the co-memory cell ${\bf c}_t$ can be expressed as 
	\begin{equation} \label{eq14}
		\begin{aligned}
			{{\bf{c }}_t} = {\bf \pi}_t^a \odot {\bf{c}}_t^a + {\bf \pi}_t^b \odot {\bf{c}}_t^b.
		\end{aligned}
	\end{equation}
	
	In the concurrent LSTM unit, two sub-memory units share a common output gate ${\bf o}_{t}$. 
	The activation of the cell gate  ${\bf o}_{t}$ is similar to the activation of the cell gate, i.e.,
	\begin{equation} \label{eq15}
		\begin{aligned}
			{{\bf{o}}_t} = \sigma ({\bf{W}}_{o{\rm{x}}}^{} \cdot \left[ {\begin{array}{*{20}{c}}
					{{\bf{x}}_t^a}\\
					{{\bf{x}}_t^b}
				\end{array}} \right] + {\bf{W}}_{oh}^{} \cdot {{\bf{h}}_{t - 1}} + {\bf{b}}_o^{}).
			\end{aligned}
		\end{equation}
		Finally, a hidden state ${{\bf{h}}_t}$ at time step $t$ can be expressed as
		\begin{equation} \label{eq16}
			\begin{aligned}
				{{\bf{h}}_t} = {{\bf{o}}_t} \odot \varphi ({{\bf{c }}_t}).
			\end{aligned}
		\end{equation}
		Briefly, at time step $t$, the proposed Co-LSTSM model proceeds in the following order. {\em
			\begin{itemize}
				\vspace{-2mm}
				\item Compute input gates ${\bf{i}}_t^s$ and forget gates ${\bf{f}}_t^s$ by Eq~\eqref{eq10} and Eq~\eqref{eq11}, respectively;
				\vspace{-2mm}
				\item Update sub-memory cells ${\bf{c}}_t^s$ by Eq~\eqref{eq13};
				\vspace{-2mm}
				\item Compute cell gates ${{\bf{\pi}}_t^s}$ by Eq~\eqref{eq13.5};
				\vspace{-2mm}
				\item Compute co-memory gate ${{\bf{c}}_t}$ by Eq~\eqref{eq14};
				\vspace{-2mm}
				\item Compute output gate ${{\bf{o}}_t}$ by Eq~\eqref{eq15};
				\vspace{-2mm}
				\item Output ${\bf h}_t$ by Eq~\eqref{eq16}.
			\end{itemize}
		}
		
		\subsection{Learning Algorithm}
		We employ a loss function to learn the model parameters of Co-LSTSM by measuring the deviation between the target class ${l_t}$ and ${\bf y}_t$ at time step $t$, i.e., $\ell ({{\bf{y}}_t},{l_t}){{ = }} - \log {y_{t,{{l}_t}}}$.
		Both types of loss functions can be minimized by Back Propagation Through Time (BPTT) algorithm~\cite{DBLP:conf/iceis/CuellarDJ05}, which unfolds the Co-LSTSM model over several time steps and then runs the back propagation algorithm to train the model. specifically, LSTM usually uses the truncated BPTT to prevent the back-propagation errors. The idea is that once the back-propagated error leaves the LSTM unit or gates, it will not be allowed to enter the LSTM unit again. Here, we also do not allow the errors to re-enter the concurrent LSTM unit once they leave the co-memory cell.

		\section{Experiments}
		\label{E}
		\subsection{Dataset}
		We conduct experiments to evaluate the performance of the proposed Co-LSTSM  by comparing with the state-of-the-art methods and some baselines on the following two widely-used benchmarks.
		
		{\bf BIT dataset~\cite{kong2012leraning}.} It consists of eight classes of human
		interactions, i.e., bow, boxing, handshake, high-five, hug, kick,
		pat, and push. In each class, there are 50 videos, which  are captured  in real scenarios  within the cluttered  backgrounds. For some videos, there are partially  occluded  bodies,  moving  objects, as well as devise appearances, scales, poses, illuminations and viewpoints. Following the setting in~\cite{kong2014interactive}, 34 videos per class are randomly chosen as training data and the remaining ones for testing.

		{\bf UT dataset~\cite{ryoo2009spatio}.}
		It consists of  ten  videos, each of which contains six classes  of  human interactions, i.e., handshake, hug, kick, point, punch and push. These videos are captured with different scales and
		illuminations. The authors provide the interaction labels for each frame.  After extracting the frames, we obtain 60 video clips in total, namely 10 video clips per class. The leave-one-out cross validation training strategy is adopted for the experiments, i.e., nine video clips per class are used for training while the remaining one for cross validation. Finally, averaged accuracy on 10 times is reported as the final performance.

		\subsection{Implementation Details}
		In the preprocessing step, the bounding box corresponding to each interacting person is  detected  and  tracked  over all frames  by
		an object detector~\cite{girshick2015fast} and object tracker~\cite{zamir2012gmcp}. Since some works validated that placing the LSTM network on fc6 of CNN performs better than fc7 of CNN~\cite{donahue2015long}, we employ the pre-trained AlexNet model~\cite{krizhevsky2012imagenet} to extract the two types of fc6 features on two bounding boxes around two concurrent people, respectively.
		
\begin{figure}[t]
			\centering
			\includegraphics[scale=0.38]{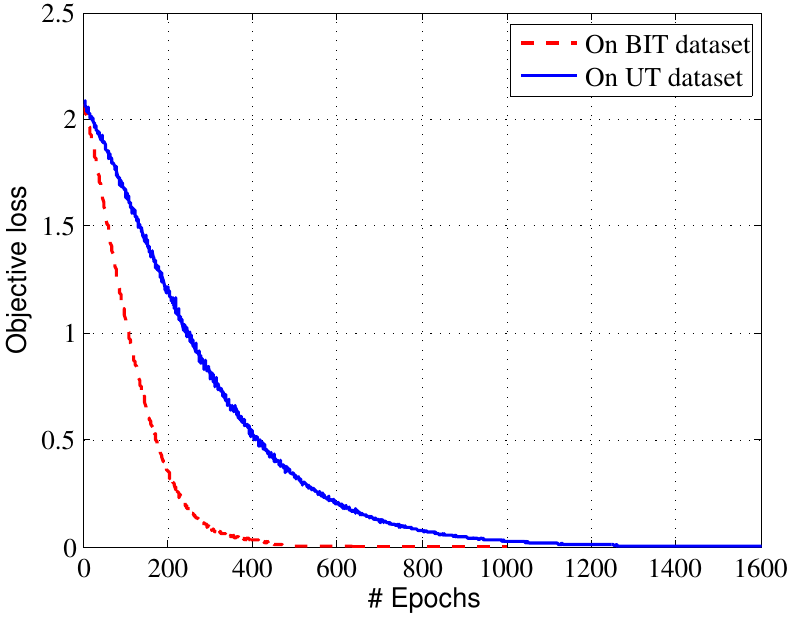}
			\caption{Objective loss curve over training epochs.}
			\vspace{-3mm}
			\label{fig_loss}
		\end{figure}

		
		For BIT dataset and UT dataset, the length $T$ of time steps is set to $30$ and $40$, respectively. 
		The sub-memory cell nodes are set 2048 on both BIT and UT. The time steps of each video clip in BIT dataset and UT dataset are set 30 and 40 respectively. We use Torch toolbox and Caffe as the deep learning platform and a NVIDIA Tesla K20 GPU to run the experiments. The learning rate, momentum and decay rate are set $1\times10^{-5}$, 0.9 and 0.95, respectively. We plot the learning curve for training Co-LSTSM model on BIT dataset and UT dataset in Figure~\ref{fig_loss}. We can see that the training of Co-LSTSM begins to converge after about $600$ and $1300$ epochs on the BIT dataset and the UT dataset, respectively.
		
				\begin{table*}[!t]
					\begin{center}
						\vspace{-2.5mm}
						\begin{tabular}{cc cc cccccc}
							\hline\hline
							\centering
							Method & bow & boxing & handshake & high-five & hug & kick & pat & push & Average \\
							\hline
							Lan {\em et al.}~\cite{lan2012discriminative} & 81.25 & 75.00 & 81.25 & 87.50 & 87.50 & 81.25 & 81.25 & 81.25 & 82.03 \\
							Liu {\em et al.}~\cite{liu2011recognizing} & 100.00 & 75.00 & 81.25 & 87.50 & 93.75 & 87.50 & 75.00 & 75.00 & 84.37 \\
							Kong {\em et al.}~\cite{kong2012leraning} & 81.25 & 81.25 & 81.25 & 93.75 & 93.75 & 81.25 & 81.25 & 87.50 & 85.16 \\
							Kong {\em et al.}~\cite{kong2016close} & 87.50 & 81.25 & 87.50 & 81.25 & 87.50 & 81.25 & 87.50 & 87.50 & 85.38 \\
							Kong {\em et al.}~\cite{kong2014interactive} & 93.75 & 87.50 & {93.75} & {93.75} & 93.75 & 87.50 & 87.50 & 87.50 & 90.63 \\
							Donahue {\em et al.}~\cite{donahue2015long} & 100.00 & 75.00 & 85.00&	69.75& 85.00& 69.75	&80.00	&76.50 &80.13 \\
							Ke {\em et al.}~\cite{ke2016spatial} & - & - &	-&	-&	-&	-&	-&	- & 85.20 \\
							\hline
							Person-box CNN    & 100.00 & 75.00& 62.50& 	56.25& 	93.75& 	68.75& 	56.25& 	62.50& 	71.88\\
							One CNN+LSTM      & 100.00 & 75.00&	84.50&	84.50&	88.00&	88.00&	70.00&	78.00&	83.50 \\
							Two CNN+LSTM      & 100.00 & 79.00& 84.50&	84.50&	{94.75}& 88.00&	80.50&	90.00&	87.66
							\\
							Co-LSTSM &100.00 & 90.50 & {92.50} & 92.50& {94.75}& {88.00} & {90.50} &{94.25} & {92.88}
							
							\\
							\hline\hline
						\end{tabular}
					\end{center}
					\caption{Recognition  accuracy (\%) of different methods on the BIT dataset.}
					\label{BIT_results}
				\end{table*}
		
		In experiments, three baselines are conducted to illustrate the novelty of the proposed Co-LSTSM.
		\begin{itemize}
			\item {\bf Person-box CNN}. The pre-trained AlexNet model is deployed on two bounding boxes around the two concurrent people at each time step respectively, where two fc6 features corresponding to two interacting people are concatenated into a long vector. Then the concatenated features over all time steps are pooled into a single feature.  All features from each video clip are trained and tested on the softmax classifier. This baseline can illustrate the importance of deep features.
			\item
			{\bf One CNN+LSTM}. This baseline treats two individual actions as a whole.   First, two bounding boxes corresponding two interacting people at each time step are merged into a bigger bounding box. Second, fc6 features are extracted by AlexNet on this ``bigger" bounding box at each time step. Third, we use the fc6 features at each time step as inputs to train a LSTM model. The model of this baseline is similar to Long-term Recurrent Convolutional Networks (LRCN)~\cite{donahue2015long}.
			\item
			{\bf Two CNN+LSTM}.  This  baseline models the individual dynamics of two people by two LSTM networks, respectively. 	First, AlexNet is deployed on the two bounding boxes around two interacting people at each time step to extract fc6 features. Second, fc6 features from two individuals are feed to one LSTM networks to capture the individual dynamics, respectively. Third, the softmax scores output from these two LSTM networks are fused. This idea of this baseline is the same as Two-Stream Convolutional Networks~\cite{simonyan2014two}.	
			
		\end{itemize}

		\begin{table*}[!t]
			\begin{center}
				\begin{tabular}{cc cc cccc}
					\hline\hline
					\centering
					Method & handshake & hug & kick & point & punch & push & Average  \\
					\hline
					Ryoo {\em et al.}~\cite{ryoo2009spatio} & 75.00 & 87.50 & 62.50 & 50.00 & 75.00 & 75.00 & 70.80  \\
					Yu {\em et al.}~\cite{yu2010real} & 100.00& 65.00& {100.00} & 85.00& 75.00& 75.00& 83.33 \\
					Ryoo ~\cite{ryoo2011human} & 80.00 & 90.00 & 90.00 & 80.00 & 90.00 & 80.00 & 85.00 \\
					Kong {\em et al.}~\cite{kong2012leraning} & 80.00 & 80.00 & {100.00} & 90.00 & 90.00 & 90.00 & 88.33 \\
					Kong {\em et al.}~\cite{kong2014interactive}  & 100.00 & 90.00 & {100.00} & 80.00 & 90.00 & 90.00 & 91.67  \\
					Kong {\em et al.}~\cite{kong2016close} & 90.00 & 100.00 & 90.00 & 100.00 & 90.00 & 90.00 & 93.33  \\
					Raptis \& Sigal~\cite{raptis2013poselet} & 100.00 & 100.00 & 90.00 & 100.00 & 80.00 & 90.00 & 93.30 \\
					Shariat \& Pavlovic~\cite{shariat2013a} & - & - & - & - & - & - & 91.57 \\
					Zhang {\em et al.}~\cite{zhang2012spatio} & 100.00 & 100.00 & {100.00}  & 90.00 & 90.00 & 90.00 & {95.00} \\
					Donahue {\em et al.}~\cite{donahue2015long} & 90.00& 80.00 & 90.00	& 80.00 & 90.00	& 80.00&  85.00 \\
					Ke {\em et al.}~\cite{ke2016spatial}  &	-&	-&	-&	-&	-&	- & 93.33\\
					Wang {\em et al.}~\cite{wang2015hierarchical}  &	-&	-&	-&	-&	-&	- & {95.00} \\
					\hline
					Person-box CNN  & 90.00 & 80.00 & 80.00 & 80.00 & 80.00 & 80.00 & 81.67   \\
					One CNN+LSTM  & 90.00	& 80.00	& 90.00	& 80.00	& 90.00	& 80.00 & 85.00\\
					TWo CNN+LSTM & 100.00	& 100.00	& 90.00	& 80.00	& 90.00	& 80.00 & 90.00\\
					{Co-LSTSM} & {100.00} &	{100.00} & {90.00}	& {100.00} & {90.00}	& {90.00} & {95.00}  \\
					\hline\hline
				\end{tabular}
			\end{center}
			\vspace{-2mm}
			\caption{Recognition  accuracy (\%) of different methods on the UT dataset.}
			\label{UT_results}
			
		\end{table*}

		\subsection{Results on the BIT dataset}
				\vspace{-1.5mm}	
		{\bf Comparison with baselines.} Table~\ref{BIT_results} shows the recognition accuracy of the proposed Co-LSTSM compared with the baselines. As shown
		in this table, Co-LSTSM significantly outperforms the baseline methods. We can see that adding the temporal information by employing LSTM (i.e., ``One CNN+LSTM", and ``Two CNN+LSTM") improves the performance of ``Person-box CNN" without temporal information. In particular, ``Two CNN+LSTM"  achieves the higher accuracy than ``One CNN+LSTM". It is illustrated that an single LSTM model can capture a single motioning object better than multiple motioning objects. 
		
		{\bf  Comparison with state-of-the-art methods.} We also compare Co-LSTSM
		with the state-of-the-art methods for person-person interaction recognition, i.e., hand-crafted spatio-temporal  interest
		points~\cite{DollarVSPETS05cuboids} based methods of Lan {\em et al.}~\cite{lan2012discriminative}, Liu {\em et al.}~\cite{liu2011recognizing}, and Kong {\em et al.}~\cite{kong2012leraning,kong2014interactive,kong2016close}, ws well as LSTM-based methods of  Donahue {\em et al.}~\cite{donahue2015long} and Ke {\em et al.}~\cite{ke2016spatial}. Table~\ref{BIT_results} lists the experimental results, in which some results are reported in~\cite{kong2014interactive,kong2016close}. We can see Co-LSTSM performs better than the comparative methods, especially all LSTM-based methods, i.e., Donahue {\em et al.}~\cite{donahue2015long} and Ke {\em et al.}~\cite{ke2016spatial}. In particular, compared
		with the state-of-the-art LSTM-based methods (i.e., Ke {\em et al.}~\cite{ke2016spatial} with 85.20\%), Co-LSTSM has gained about 8\% improvement.

		
		
		
		\subsection{Results on the UT dataset}
				\vspace{-1.5mm}	
		{\bf Comparison with baselines.} Table~\ref{UT_results} shows the recognition accuracy of the proposed Co-LSTSM compared with the baselines. It is observed that Co-LSTSM  performs consistently better than all baselines. ``One CNN+LSTM" and ``Two CNN+LSTM" considering the temporal information performs better than ``Person-box CNN" without temporal information. In particular, ``Two CNN+LSTM"  achieves the better performance than ``One CNN+LSTM". 
		
		{\bf Comparison with state-of-the-art methods.} Co-LSTSM is also compared
		with the state-of-the-art methods, including some traditional methods (i.e., Ryoo {\em et al.}~\cite{ryoo2009spatio}, Yu {\em et al.}~\cite{yu2010real}, Kong {\em et al.}~\cite{kong2012leraning,kong2014interactive,kong2016close},
		Raptis \& Sigal~\cite{raptis2013poselet}, Shariat \& Pavlovic~\cite{shariat2013a}, and Zhang {\em et al.}~\cite{zhang2012spatio}), deep learning method (i.e., Wang {\em et al.}~\cite{wang2015hierarchical}), as well as LSTM-based methods (i.e., Ke {\em et al.}~\cite{ke2016spatial} and Donahue {\em et al.}~\cite{donahue2015long}).
		The comparison results are shown in Table~\ref{UT_results}. We can see that Co-LSTSM also achieves the state-of-the-arts result, i.e., 95.00\% by Zhang {\em et al.}~\cite{zhang2012spatio} and Wang {\em et al.}~\cite{wang2015hierarchical}. It is noted that Wang {\em et al.}~\cite{wang2015hierarchical} adopted deep context features on the event neighborhood, where the size of event neighborhood need be manually defined in the preprocessing step; Zhang {\em et al.} proposed a spatio-temporal phrase to capture a certain number of local movements between interacting people, where the number of local movements increases when the interaction becomes complex. As new exploration by leveraging LSTM model, the proposed Co-LSTSM performs better than other LSTM-based methods, i.e., Donahue {\em et al.}~\cite{donahue2015long} and Ke {\em et al.}~\cite{ke2016spatial}.

		
		\subsection{Evaluation on Human Interaction Prediction}
				\vspace{-2.0mm}	
		In this work, we also evaluate the proposed Co-LSTSM on human interaction prediction. Unlike person-person interaction recognition, human interaction prediction is defined to recognize an ongoing interaction activity before the interaction is completely executed~\cite{ke2016spatial}. Due to the large variations in appearance and the evolution of scenes, interaction prediction at an early stage is a challenging task. Following experimental setting in~\cite{ke2016spatial,kong2014max}, a testing video clip is divided into 10 incomplete action executions by using 10 observation ratios (i.e., from 0 to 1 with a step size of 0.1), which represent the increasing amount of sequential data with time. For example, given a testing video clip with the length $T$, a prediction accuracy under an observation ratio of $0.3$ denotes that the accuracy is tested with the first length $0.3\times T$ frames. When the observation ratio is $1$, namely the
		entire video clip is used, Co-LSTSM acts as the person-person interaction recognition model.
		
\begin{figure}[t]
			\centering
			\includegraphics[scale=0.47]{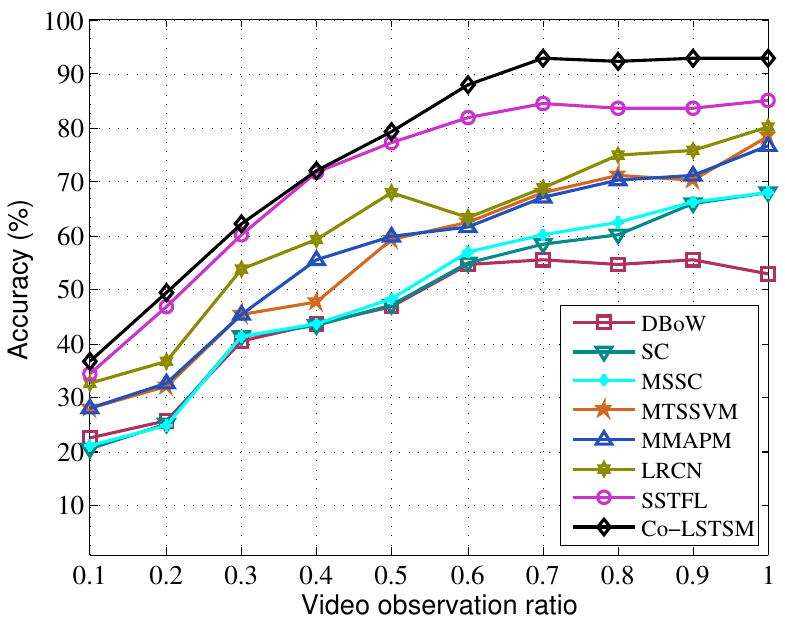}
			\vspace{-2mm}
			\caption{Performance of human interaction prediction on BIT.}
			\label{fig_prediction}
			\vspace{-2mm}
		\end{figure}

		
		The comparative methods includes Dynamic
		Bag-of-Words (DBoW)~\cite{ryoo2011human}, Sparse Coding (SC)~\cite{cao2013recognize}, Sparse Coding with Mixture
		of training video Segments (MSSC)~\cite{cao2013recognize}, Multiple Temporal Scales based on SVM (MTSSVM)~\cite{kong2014a}, Max-Margin Action Prediction Machine (MMAPM)~\cite{kong2014max}, Long-term Recurrent Convolutional Networks (LRCN)~\cite{donahue2015long}, and Spatial-Structural-Temporal Feature Learning (SSTFL)~\cite{ke2016spatial}. The comparison results on the BIT dataset with different observation ratios are listed
		in Figure~\ref{fig_prediction}. Overall, Co-LSTSM outperforms all comparative methods for all observation ratios. Specifically, we can see that 1) the improvement of Co-LSTSM is more significant when the observation ratio is $0.6$; 2) the accuracy of Co-LSTSM increases rapidly when the observation ratio is $0.5$, which illustrates the close interaction is happening; and 3) the accuracy of Co-LSTSM becomes stable when the observation ratio is $0.7$, which illustrates the close interaction is ending.

		\section{Conclusions and Future Work}
				\vspace{-2.5mm}	
		In this work, for person-person interaction recognition, we propose a novel Concurrence-Aware Long Short-Term Sub-Memories (Co-LSTSM) to aggregate the interactive motions between interacting people over time. Specifically, interacting people at each time step are jointly modeled by a novel {concurrent} LSTM unit, which captures the concurrently inter-related motion information from two sub-memory units. Experimental results on person-person interaction recognition and prediction have demonstrated the superior performance of the proposed Co-LSTSM compared with the state-of-the-art methods. In future, we will extend Co-LSTSM for addressing the problem of complex group collective activity analysis.

		{\small
			\bibliographystyle{ieee}
			\bibliography{egbib}
		}
\end{document}